\documentclass{article}

\usepackage{arxiv}
\usepackage{caption}
\usepackage{subcaption}
\usepackage{graphicx}
\usepackage[utf8]{inputenc} 
\usepackage[T1]{fontenc}    
\usepackage{hyperref}       
\usepackage{url}            
\usepackage{booktabs}       
\usepackage{amsfonts}       
\usepackage{amsmath}
\usepackage{nicefrac}       
\usepackage{microtype}      
\usepackage{lipsum}
\DeclareMathOperator*{\argmax}{arg\,max}

\title{Self-attention based end-to-end Hindi-English Neural Machine Translation}

\author{Siddhant Srivastava \\
  Soft Computing Laboratory \\
  ABV-IIITM Gwalior \\
  {\tt siddhant.srivastava11@gmail.com} \\\And
  Ritu Tiwari \\
  Soft Computing Laboratory \\
  ABV-IIITM Gwalior \\
  {\tt tiwariritu2@gmail.com}}

\begin{document}
\maketitle

\begin{abstract}
 Machine Translation (MT) is a zone of concentrate in Natural Language processing which manages the programmed interpretation of human language, starting with one language then onto the next by the PC. Having a rich research history spreading over about three decades, Machine interpretation is a standout amongst the most looked for after region of research in the computational linguistics network. As a piece of this current ace's proposal, the fundamental center is examine the Deep-learning based strategies that have gained critical ground as of late and turning into the \textit{de facto} strategy in MT. We would like to point out the recent advances that have been put forward in the field of Neural Translation models, different domains under which NMT has replaced conventional SMT models and would also like to mention future avenues in the field. Consequently, we propose an end-to-end self-attention transformer network for Neural Machine Translation, trained on Hindi-English parallel corpus and compare the model's efficiency with other state of art models like encoder-decoder and attention based encoder-decoder neural models on the basis of BLEU. We conclude this paper with a comparitive analysis of the three proposed models.   
\end{abstract}


\section{Introduction}
Machine Translation, which is a field of concentrate under common language preparing, focuses at deciphering normal language naturally utilizing machines. Information driven machine interpretation has turned into the overwhelming field of concentrate because of the availability of substantial parallel corpora. The primary target of information driven machine interpretation is to decipher concealed source language, given that the frameworks take in interpretation learning from sentence adjusted bi-lingual preparing information. 

Statistical Machine Translation (SMT) is an information driven methodology which utilizes probabilistic models to catch the interpretation procedure. Early models in SMT depended on generative models accepting a word as the fundamental element \cite{Brown_1903}, greatest entropy based discriminative models utilizing highlights gained from sentences \cite{Och_2002}, straightforward and various leveled phrases \cite{Koehn_2003,Chiang_2017}. These strategies have been extraordinarily utilized since 2002 regardless of the way that discriminative models looked with the test of information sparsity. Discrete word based portrayals made SMT vulnerable to learning poor gauge on the record of low check occasions. Additionally, structuring highlights for SMT physically is a troublesome errand and require area language, which is hard remembering the assortment and intricacy of various common dialects.

Later years have seen the extraordinary accomplishment of deep learning applications in machine interpretation. Deep learning approaches have surpassed factual strategies in practically all sub-fields of MT and have turned into the \textit{de facto} technique in both scholarly world just as in the business. as a major aspect of this theory, we will talk about the two spaces where deep learning has been significantly utilized in MT. We will quickly examine \textit{Component or Domain based deep learning strategies for machine translation} \cite{Devlin_2014} which utilizes deep learning models to improve the viability of various parts utilized in SMT including language models, transition models, and re-organizing models. Our primary spotlight in on \textit{end-to-end deep learning models for machine translation} \cite{Sutskever_2014,Bahdanau_2014} that utilizes neural systems to separate correspondence between a source and target language straightforwardly in an all encompassing way without utilizing any hand-created highlights. These models are currently perceived as \textit{Neural Machine translation} (NMT).

Let $x$ signify the source language and $y$ mean the objective language, given a lot of model parameters $\theta$ , the point of any machine interpretation calculation is to discover the interpretation having greatest likelihood $\hat{y}$:
\begin{equation}
    \hat{y} = \argmax_y {P(y|x;\theta)}.
\end{equation}
The decision rule is re-written using Bayes' rule as \cite{Brown_1903}:
\begin{equation}
    \hat{y} = \argmax_y {\frac{P(y;\theta_{lm})P(x|y;\theta_{tm})}{P(x)}}.
    \end{equation}
\begin{equation}
   \hat{y} = \argmax_y {P(y;\theta_{lm})P(x|y;\theta_{tm})}. 
\end{equation}
Where $P(y;\theta_{lm})$ is called as \textit{language model}, and $P(x|y;\theta_{tm})$ is called as \textit{transition model}. 
The interpretation model likewise, is characterized as generative model, which is crumbled by means of dormant structures. 
\begin{equation}
    P(x|y;\theta_{tm}) = \sum_{z}P(x,y|z;\theta_{tm}).
\end{equation}
Where, $z$ signifies the idle structures like word arrangement between source language and target language.

\section{End-to-End Deep Learning for Machine translation} 

Start to finish Machine Translation models \cite{Sutskever_2014,Bahdanau_2014} likewise named as Neural Machine Translation (NMT), intends to discover a correspondence among source and target normal dialects with the assistance of deep neural systems. The fundamental distinction among NMT and customary Statistical Machine Translation (SMT) \cite{Brown_1903,Vogel_1996,Koehn_2003,Och_2002} based methodologies is that Neural model are fit for learning complex connections among characteristic dialects straightforwardly from the information, without turning to manual hand highlights, which are difficult to plan. 

The standard issue in Machine Translation continues as before, given an arrangement of words in source language sentence $X = x_{1},....x_{j},....x_{J}$ and target language sentence $Y = y_{1},....y_{i},....y_{I}$, NMT endeavors to factor sentence level interpretation likelihood into setting dependant sub-word interpretation probabilities. 

\begin{equation}
    P(y|x;\theta) = \prod_{i=1}^{I} P(y_{i}|x,y_{<i};\theta)
\end{equation}

Here $y_{<i}$ is alluded to as fractional interpretation. There can be sparsity among setting among source and target sentence when the sentences become excessively long, to tackle this issue, \cite{Sutskever_2014} proposed an encoder-decoder arrange which could speak to variable length sentence to a fixed length vector portrayal and utilize this conveyed vector to decipher sentences.

\subsection{Encoder Decoder Framework for Machine Translation} 

\begin{figure}
\centering
\includegraphics[width=0.6\textwidth]{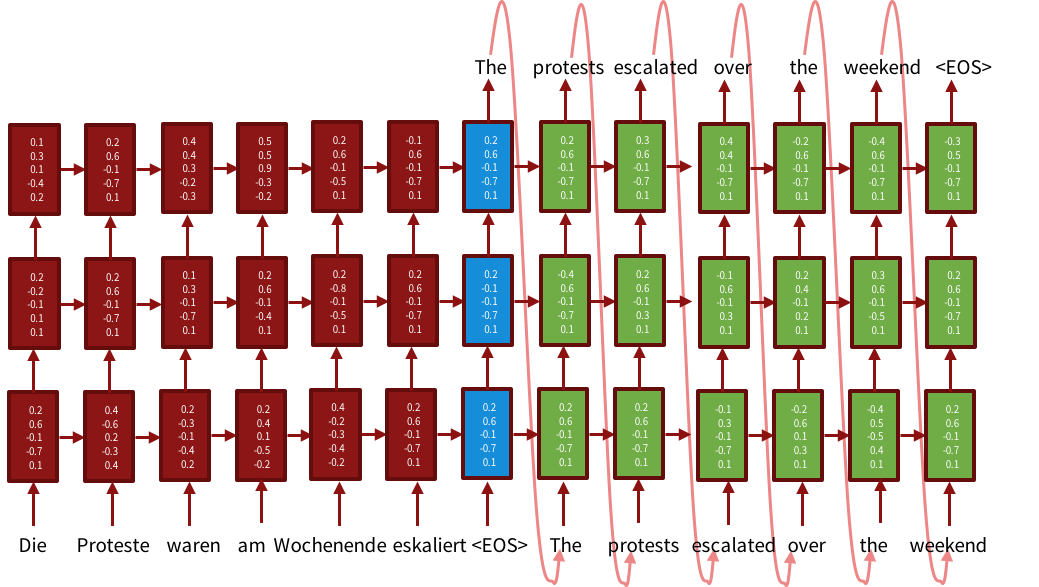}
\caption{Encoder-Decoder model for Machine Translation, Crimson boxes portray the concealed expressed of encoder, Blue boxes indicates "End of Sentence" EOS and Green boxes show shrouded condition of the decoder. credits (Neural Machine Translation - Tutorial ACL 2016)}
\label{fig:encdec}
\end{figure}

Neural Machine Translation models stick to an Encoder-Decoder engineering, the job of encoder is to speak to subjective length sentences to a fixed length genuine vector which is named as setting vector. This setting vector contains all the fundamental highlights which can be construed from the source sentence itself. The decoder arrange accepts this vector as contribution to yield target sentence word by word. The perfect decoder is relied upon to yield sentence which contains the full setting of source language sentence. Figure\ref{fig:encdec} shows the overall architecture of the encoder-decoder neural network for machine translation.

Since source and target sentences are ordinarily of various lengths, Initially \cite{Sutskever_2014} proposed Recurrent Neural Network for both encoder and decoder systems, To address the issue of evaporating angle and detonating slopes happening because of conditions among word sets, Long Short Term Memory (LSTM) \cite{Hochreiter_1997} and Gated Recurrent Unit (GRU) \cite{Cho_2014} were proposed rather than Vanilla RNN cell.

Training in NMT is done by maximising log-likelihood as the objective function:
\begin{equation}
    \hat{\theta} = \argmax_\theta L(\theta)
\end{equation}
Where $L(\theta)$ is defined as:
\begin{equation}
    L(\theta) = \sum_{i=1}^{I} log P(y^{(i)}|x^{(i)};\theta)
\end{equation}
After training, learned parameters $\hat{\theta}$ is used for translation as: 
\begin{equation}
    \hat{y} = \argmax_y P(y|x;\hat{\theta})
\end{equation}

\subsection{Attention Mechanism in Neural Machine Translation}

The Encoder organize proposed by \cite{Sutskever_2014} spoke to source language sentence into a fixed length vector which was in this way used by the Decoder arrange, through observational testing, it was seen that the nature of interpretation incredibly relied upon the span of source sentence and diminished essentially by expanding the sentence measure. 

To address this issue, \cite{Bahdanau_2014} proposed to coordinate an Attention system inside the Encoder arrange and demonstrated this could progressively choose significant parts of setting in source sentence to deliver target sentence. They utilized Bi-directional RNN (BRNN's) to catch worldwide settings:
\begin{equation}
    \overrightarrow{h_{s}} = f(x_{(s)},\overrightarrow{h_{s-1}},\theta)
\end{equation}

\begin{equation}
    \overleftarrow{h_{s}} = f(x_{(s)},\overleftarrow{h_{s-1}},\theta)
\end{equation}
The forward hidden state $\overrightarrow{h_{s}}$ and backward hidden state $\overleftarrow{h_{s}}$ are concatenated to capture sentence level context.
\begin{equation}
    h_{s} = [\overrightarrow{h_{s-1}};\overleftarrow{h_{s-1}}]
\end{equation}

The basic Ideology behind computing \textit{attention} is to seek  portions of interest in source text in order to generate target words in text, this is performed by computing attention weights first.
\begin{equation}
    \alpha_{j,i} = \frac{exp(a(t_{j-1},h_{i},\theta))}{\sum_{i'=1}^{I+1}exp(a(t_{j-1},h_{i'},\theta))}
\end{equation}

\begin{figure} [t]
\centering
\includegraphics[width=0.5\textwidth]{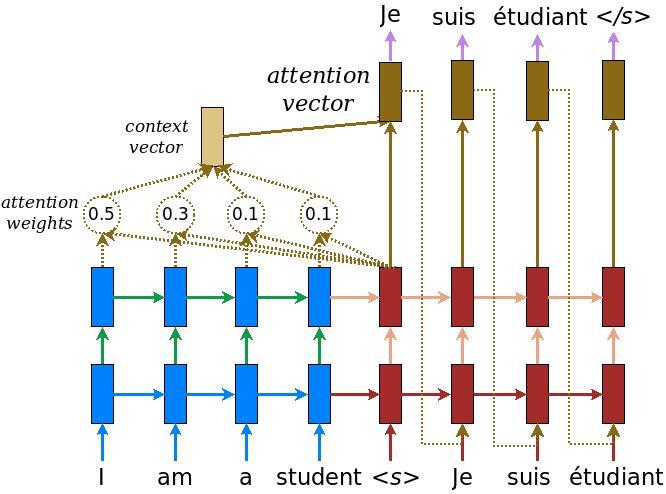}
\caption{Consideration based Encoder-Decoder Architecture for Machine Translation. The majority of the Architecture is like fundamental Encoder-Decoder with the expansion of Context Vector Computed utilizing consideration loads for each word token, Attention vector is determined utilizing Context vector and concealed condition of encoder. credits (Attention-based Neural Machine Translation with Keras, blog by Sigrid Keydana)}
\label{fig:attmech}
\end{figure}

Where $a(t_{j-1},h_{i},\theta)$ is the alignment function which evaluates how well inputs are aligned with respect to position $i$ and output at position $i$. Context vector $c_{j}$ is computed as a weighted sum of hidden states of the source.

\begin{equation}
    c_{j} = \sum_{i=1}^{I+1} \alpha_{j,i}h_{i}
\end{equation}
And target hidden state is computed as follows. 
\begin{equation}
    t_{j} = f(y_{j-1},s_{j-1},c_{j},\theta)
\end{equation}
In Figure \ref{fig:attmech}, we have attention mechanism at the encoder level, the context vector is then used by the decoder layer for language translation. The distinction between consideration based NMT \cite{Bahdanau_2014} from unique encoder-decoder based engineering \cite{Sutskever_2014} is the way source setting is registered, in unique encoder-decoder, the source's shrouded state is utilized to introduce target's underlying concealed state while in consideration instrument, a weighted aggregate of concealed state is utilized which ensures that the significance of every single source word in the sentence is very much protected in the specific circumstance. This incredibly improves the execution of interpretation and hence this has turned into the \textit{state of art} model in neural machine interpretation.

\section{Neural Architectures for NMT}

The majority of the encoder-decoder based NMT models have used RNN and It's variations LSTM \cite{Hochreiter_1997} and GRU \cite{Cho_2014}. As of late, Convolution systems (CNN) \cite{Gehring_2017} and self consideration systems \cite{Vaswani_2017} have been examined and have delivered promising outcomes. 

The issue with utilizing Recurrent systems in NMT is that it works by sequential calculation and necessities to keep up it's concealed advance at each progression of preparing. This makes the preparation deeply wasteful and tedious. \cite{Gehring_2017} proposed that convolution systems can, interestingly, become familiar with the fixed length shrouded states utilizing convolution task. The principle preferred standpoint of this methodology being that convolution task doesn't rely upon recently figured qualities and can be parallelized for multi-center preparing. Additionally Convolution systems can be stacked in a steady progression to learn further setting settling on it a perfect decision for both the encoder and decoder. 

Intermittent systems process reliance among words in a sentence in $O(n)$ while Convolution system can accomplish the equivalent in $O(log_{k}n)$ where $k$ is the extent of convolution part. 

\begin{figure} [t]
\centering
\includegraphics[width=0.4\textwidth]{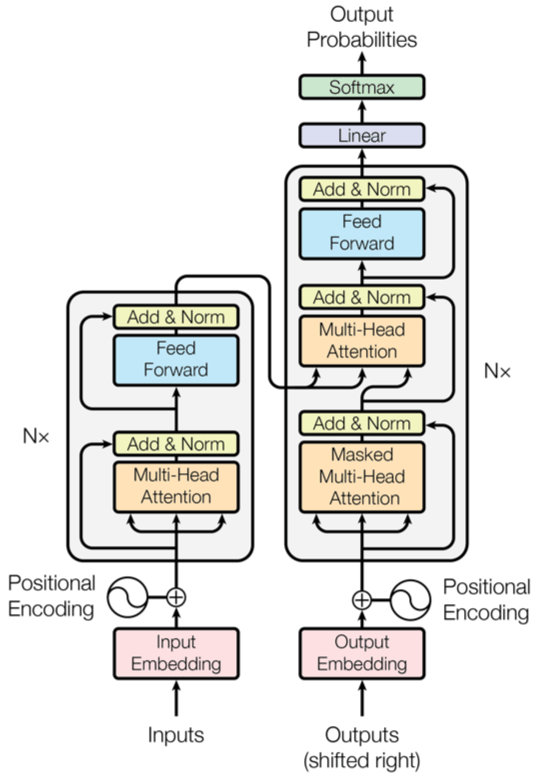}
\caption{Self-Attention Encoder-Decoder Transformer model. Encoder and Decoder both consists positional encoding and stacked layers of multi-head attention and feedforward network with the Decoder containing an additional Masked multi head attention. Transition Probabilities are calculated using linear layer followed by softmax. credits (Vaswani et al. 2017)}
\label{fig:model}
\end{figure}

\cite{Vaswani_2017} proposed a model which could register the reliance among each word pair in a sentence utilizing just the Attention layer stacked in a steady progression in both the encoder and decoder, he named this as \textit{self-attention}, the overall architecture is shown as Figure \ref{fig:model}. In their model, concealed state is figured utilizing self-consideration and feed forward system, they utilize positional encoding to present the element dependent on the area of word in the sentence and their self-consideration layer named as \textit{multi-head attention} is very parallelizable. This model has appeared to be exceedingly parallelizable due to before referenced reason and fundamentally accelerates NMT preparing, likewise bringing about preferable outcomes over the benchmark Recurrent system based models.

As of now, there is no clear decision regarding which neural architecture is the best and different architectures give different results depending on the problem in hand. Neural architecture is still considered to be the hottest and most active research field in Neural Machine Translation.

\section{Research gaps and open problems} 

deep learning strategies have altered the field of Machine Translation, with early endeavors concentrating on improving the key segments of Statistical Machine Translation like word arrangement \cite{Yang_2013} , interpretation model \cite{Koehn_2003,Gao_2014}, and expression reordering \cite{Li_2013,Li_2014} and language model \cite{Vaswani_2003}. Since 2010, a large portion of the exploration has been moved towards creating start to finish neural models that could relieve the need of broad component designing \cite{Sutskever_2014,Bahdanau_2014}. Neural models have effectively supplanted Statistical models since their commencement in all scholarly and modern application. 

Albeit Deep learning has quickened look into in Machine Translation people group yet regardless, Current NMT models are not free from blemishes and has certain constraints. In this segment, we will depict some current research issues in NMT, our point is to control specialists and researchers working in this field to get to know these issues and work towards it for considerably quicker improvement in the field. 

\subsection{Neural models motivated by semantic approaches} 

Start to finish models have been named as the \textit{de facto} model in Machine Translation, yet it is difficult to decipher the inner calculation of neural systems which is frequently essentially said to be the \textit{"Black Box"} approach. One conceivable zone of research is to grow etymologically propelled neural models for better interpretability. It is difficult to perceive learning from concealed condition of current neural systems and thus it is similarly hard to join earlier information which is emblematic in nature into consistent portrayal of these states \cite{Ding_2017}. 

\subsection{Light weight neural models for learning through inadequate data} 

Another real disadvantage for NMT is information shortage, It is surely known that NMT models are information hungry and requires a great many preparing cases for giving best outcomes. The issue emerges when there isn't sufficient parallel corpora present for the majority of the language matches on the planet. In this way fabricating models that can adapt better than average portrayal utilizing generally littler informational collection is an effectively inquired about issue today. One comparative issue is to create one-to-numerous and many-to-numerous language models rather than balanced models. Analysts don't know how to normal information utilizing neural system from an etymological point of view, as this learning will help create multi-lingual interpretation models rather than balanced models utilized today. 

\subsection{Multi-modular Neural Architectures for present data} 

One more issue is to create multi-modular language interpretation models. Practically all the work done has been founded on printed information. Research on creating nonstop portrayal combining content, discourse and visual information to create multi-model frameworks is going all out. Additionally since there is constrained or no multi-model parallel corpora present, advancement of such databases is likewise a fascinating field to investigate and can likewise profit multi-modular neural designs. 

\subsection{Parallel and conveyed calculations for preparing neural models} 

At long last, current neural designs depend intensely broad calculation control for giving skillful outcomes \cite{Gilvile_2017_j,Castilho_2018_j, Karakanta_2018_j}. In spite of the fact that there is no figure and capacity lack in current situation, yet it would be increasingly proficient to thought of light neural models of language interpretation. Additionally Recurrent models \cite{Sutskever_2014,Bahdanau_2014} can't be parallelized because of which it is difficult to create conveyed frameworks for model preparing. Luckily, late advancements, with the rise of Convolution systems and self-consideration Networks can be parallelized and therefore disseminated among various frameworks. But since they contain a great many related parameters, it makes it difficult to circulate them among inexactly coupled frameworks. Along these lines growing light neural designs intended to be circulated can be new likely wilderness of NMT.

\section{Methodology}
The proposed methodology can be broken down to several atomic objectives. The first step is the Acquisition of parallel corpora, the next step is to pre-process the data acquired. Various neural models is to be implemented and trained on the pre-processed data. The last part of our study is to compare the results obtained by the models and do a comparative study.

\subsection{Data Acquisition and preparation}
For this study, we intend to work with is the English-Hindi parallel corpus, curated and made publically available by the Center of Indian Language Technologies (CFILT), Indian Institute of Technology, Bombay \cite{Kunchukuttan_2017}. Table \ref{tab:table2} shows the number of parallel sentences in the train and test data. This parallel datasets contains more than 1.5 million parallel sentences for training and testing purpose, to the best our knowledge, there is no literature present till date indicating any comparative study done based upon the Neural models on this dataset. 

\begin{table}[]
\caption{Statistics of Dataset used}
\label{tab:table2}
\centering
\begin{tabular}{|l|l|}
\hline
{\textbf{Dataset}}    & \textbf{Sentence Pairs}   \\
\hline
\textbf{IITB-CFILT Hi-En} &  1,495,847                         \\
\hline
Train & 1,492,827                 \\
\hline
Validation-Test           & 3,020                   \\ 
\hline
\end{tabular}
\end{table}
After getting our data in an unzipped form, the next part in our pipeline is to decompose rare words in our corpora using subword byte pair encoding \cite{Sennrich_2015}. Byte pair encoding is a useful approach when we have an extremely large vocabulary which hinders model training and thus we can decompose those rare words into common subwords and build the vocabulary accordingly.To encode the training corpora using BPE, we need to generate BPE operations first. The following command will create a file named bpe32k, which contains 32k BPE operations.It also outputs two dictionaries named vocab.de and vocab.en. Similar methodology is applied for Hindi-english data as well.

\section{Model components}
For this study, sequence-to-sequence LSTM network and Attention based encoder-decoder using GRU cell have been implemented. Self attention Transformer network has been implemented and all the models are tested side by side to create a clear superiority distinction among them. The basic theory of model components used is given below.\\
\subsection{RNN Cell}
The basic neural cell present in Neural network works well for several problems but fails miserably when the order of data matters, as a result these models fails to generalize and solve problems which deals with temporal or sequential data. To reason behind this failure being that the basic neural cell doesn't take into account the previous or backward information for it's computation and using the same philosophy the basic RNN cell was developed. Recurrent Neural Network (RNN) are such network having recurrent cells which are capable of incorporating past information with current information in terms of value computation and as a result these models have seen huge success in problems dealing with sequential input like problems coming under the domain of Natural Language Processing, weather forecast and other such problems.\\

The basic mathematical equations underlying the RNN cell are Described below:\\
\begin{equation}
    h_{t} = f_{W}(h_{t-1},x_{t})
\end{equation}
\begin{equation}
    h_{t} = tanh(W_{hh}h_{t-1},W_{xh}x_{t})
\end{equation}
\begin{equation}
    y_{t} = W_{hy}h_{t}
\end{equation}

Here $x_{t}$ and $y_{t}$ are the input and output at the $t^{th}$ time step, $W_{hh}$, $W_{xh}$ and $W_{hy}$ are connection weights respectively.

\subsection{GRU Cell}
Although the RNN cell has outperformed non-sequential neural networks but they fail to generalize to problems having longer sequence length, the problem arises due to not able to capture long term dependencies among the sequential units and this phenomena is termed as the vanishing gradient problem, To solve this problem \cite{Cho_2014} proposed a Gated approach to explicitly caputure long term memory using different cells, this cell was termed as Gated Recurrent Unit (GRU). The schematic diagram of GRU cell is given in Figure \ref{fig:awesome_image1}.

The difference between GRU cell and RNN cell lies at the computation of Hidden cell values, GRU uses two gates update ($z$) and reset ($r$) to capture long term dependancies. The mathematical equations behind the computation are given below.
\begin{equation}
    z_{t} = \sigma_{g}(W_{z}x_{t} + U_{z}h_{t-1} + b_{z})
\end{equation}
\begin{equation}
    r_{t} = \sigma_{g}(W_{r}x_{t} + U_{r}h_{t-1} + b_{r})
\end{equation}
\begin{equation}
    h_{t} = z_{t}\odot h_{t-1} + (1-z_{t})\odot\sigma_{h}(W_{h}x_{t} + U_{h}(r_{t}\odot x_{t}) + b_{h})
\end{equation}

\subsection{LSTM Cell}
\begin{figure}[ht] 
  \begin{minipage}[b]{0.5\linewidth}
    \centering
    \includegraphics[width=0.8\textwidth]{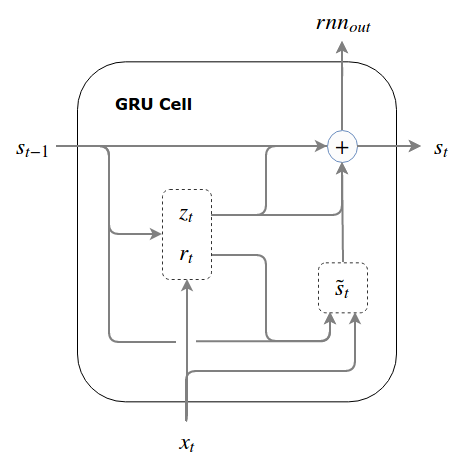}\label{fig:GRU}
\subcaption{GRU cell}
 \vspace{4ex}
  \end{minipage}
  \begin{minipage}[b]{0.5\linewidth}
    \centering
    \includegraphics[width=0.9\textwidth]{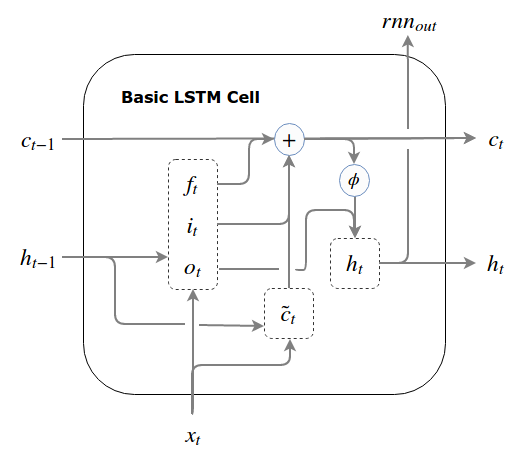}\label{fig:LSTM}
    \subcaption{LSTM}
    \vspace{4ex}
  \end{minipage}
  \caption{Cell Structure}%
  \label{fig:awesome_image1}
  \end{figure}

Short for Long Short Term Memory, Given by \cite{Hochreiter_1997} is another approach to overcome the vanishing gradient problem in RNN, like GRU, LSTM uses gating mechanism but it uses three gates instead of two cells in GRU to capture long Term Dependencies. The schematic diagram of LSTM cell is given in Figure \ref{fig:awesome_image1}.

LSTM cell uses the input ($i$), output ($o$) and forget ($f$) gates for computation of hidden states respectively, the equations are similar to that of GRU cell, LSTM like GRU, uses sigmoid activation for adding non-linearity to the function.

\begin{equation}
    f_{t} = \sigma_{g}(W_{f}x_{t} + U_{f}h_{t-1} + b_{f})
\end{equation}
\begin{equation}
    i_{t} = \sigma_{g}(W_{i}x_{t} + U_{i}h_{t-1} + b_{i})
\end{equation}
\begin{equation}
    o_{t} = \sigma_{g}(W_{o}x_{t} + U_{o}h_{t-1} + b_{o})
\end{equation}
\begin{equation}
    c_{t} = f_{t} \odot c_{t-1} + i_{t}\odot\sigma_{c}(W_{c}x_{t} + U_{c}h_{t-1} + b_{c})
\end{equation}
\begin{equation}
    h_{t} = o_{t} \odot \sigma_{c}(c_{t})
\end{equation}
\subsection{Attention Mechanism}
Attention mechanism was first developed by \cite{Bahdanau_2014}, in their paper “Neural Machine Translation by Jointly Learning to Align and Translate” which takes in as a natural extension of their previous work on the sequence to sequence Encoder-Decoder model.
Attention is proposed as a solution to mitigate the limitation of the Encoder-Decoder architecture which encodes the input sequence to one fixed length vector from which the output is decoded at each time step. This problem seems to be more of a issue when decoding long sequences.
Attention is proposed as a singular method to both align and translate.
Alignment is the problem in machine translation that seeks to find which parts of the input sequence are relevant to each word in the output, whereas translation is the process of using the relevant information to select the appropriate output.

\subsection{Transformer Network}

We use transformer self-attention encoder for our study. The transformer model \cite{Vaswani_2017} is made up of $M$ consecutive blocks. Each block of the transformer, denoted by $transformer_{l}$, contains two separate components, multi-head attention and a feed forward network. The output of each token $j$ of block $l$ is connected to it's input in a residual connection. The input to the first block is $b_{j}^{0} = x_{j}$.
\begin{equation}
    b_{j}^{l} = b_{j}^{l-1} + Transformer_{l}(b_{j}^{l-1})
\end{equation}
Multi-head attention applies self-attention over the same inputs multiple times by using separately normalized parameters (attention heads) and finally concatenates the results of each head, multi-head attention mechansim is considered as a better alternative to applying one pass of attention with more parameters as the former process can be easily parallelized. Furthermore, computing the attention with multiple heads make it easier for the
model to learn and attend to different types of relevant information with each head. The self-attention updates input $b_{j}^{l-1}$ by computing a weighted sum over all tokens in the sequence, weighted by their importance for modeling token $j$.

Each input, inside the multi-head attention is projected to query, key and value ($q,k,v$) respectively. $q,k$ and $v$ are all of dimensions $\in\mathbb{R}^{d/H}$, where $H$ is the number of heads and $d$ is the dimension of embedding. the attention weights $a_{mnh}$ for head $h$ between token $m$ and $n$ is given by scaled dot product between
\begin{equation}
    a_{mnh} = \sigma(\frac{q_{mh}^{T}k_{nh}}{\sqrt{d}})
\end{equation}
\begin{equation}
    o_{mh} = \sum_{j} v_{jh} \odot a_{mjh} 
\end{equation}

Finally the output of each head in multi-head attention in concatenated serially.

\section{Experiments and results}
\subsection{Neural models}
For this study, self attention based transformer network is implemented and compared using Sequence-to-sequence and attention based encoder decoder neural architectures. All the implementation and coding part is done using the above mentioned programming framework. We train all the three models in an end to end manner using CFILT Hindi-English parallel corpora and the results from all the three models are compared keeping in mind the usage of similar hyper-parameter values for ease of comparison.

For the sequence-to-sequence model, we are using LSTM cell for computation since there is no attention mechanism involved and it is desirable for the model to capture long term dependencies. Since LSTM works far better than GRU cell in terms of capturing long term dependencies we chose to go with it. The embedding layer, hidden layer is taken to be of 512 dimension, both encoder and the decoder part contains 2 as the number of hidden layers. For regularization, we are using dropout and set the value to be 20 percent. Batch size of 128 is taken.

For the attention based RNN search, we are using GRU cell for computation since attention mechanism is already employed and will capture long term dependencies explicitly using attention value. GRU cell is computationally efficient in terms of computation as compared with LSTM cell. Like in our sequence-to-sequence model, the embedding layer, hidden layer is taken to be of 512 dimension, both encoder and the decoder part contains 2 as the number of hidden layers. For regularization, we are using dropout and set the value to be 20 percent. Batch size of 128 is taken.
For the self attention Transformer network, we are using hidden and embedding layer to be of size 512, for each encoder and decoder, we fix the number of layers of self-attention to be 4. In each layer, we assign 8 parallel attention heads and the hidden size of feed forward neural network is taken to be 1024 in each cell. attention dropout and residual dropout is taken to be 10 percent. Table \ref{tab:table3}: shows the number of trainable parameters in each of our three models.

\begin{table*}[]
\caption{Number of Trainable parameters in each model}
\label{tab:table3}
\centering
\begin{tabular}{|l|l|}
\hline
{\textbf{Model}}    & \textbf{Trainable parameters}   \\
\hline
Sequence-to-Sequence        & 38,678,595                \\
\hline
Attention Encoder-Decoder                    & 38,804,547                    \\
\hline
Self-Attention Transformer           & 122,699,776                  \\
\hline
\end{tabular}
\end{table*}

The optimizer used for our study is the Adam optimizer, taking learning\_rate decay value of 0.001, $\beta1$ value of 0.9 and $\beta2$ value of 0.98 for first order and second order gradient moments respectively. The accuracy metric taken is log-loss error between the predicted and actual sentence word. We are trying to optimize the log-loss error between the predicted and target words of the sentence. All the models are trained on 100000 steps, where is each step, a batch size of 128 is taken for calculating the loss function. The primary objective of our training is to minimize the log-loss error between the source and target sentences and simultaneously maximize the metric which is chosen to be the BLEU score \cite{Papineni_2002}.

\section{Results}
Table \ref{tab:table4} shows the BLEU score of all three models based on English-Hindi, Hindi-English on CFILT's test dataset respectively.
From the results which we get, it is evident that the transformer model achieves higher BLEU score than both Attention encoder-decoder and sequence-sequence model. Attention encoder-decoder achieves better BLEU score and sequence-sequence model performs the worst out of the three which further consolidates the point that if we are dealing with long source and target sentences then attention mechanism is very much required to capture long term dependencies and we can solely rely on the attention mechanism, overthrowing recurrent cells completely for the machine translation task.

\begin{table*}[]
\caption{Model performance in terms of BLEU score on the task of Hindi-English and English-Hindi translation task}
\label{tab:table4}
\centering
\begin{tabular}{|l|l|l|}
\hline
{\textbf{Model}}    & \textbf{Hindi-English}   & \textbf{English-Hindi}\\
\hline
Sequence-to-Sequence        & 9.40        &       8.38 \\
\hline
Attention Encoder-Decoder                    & 11.59       &  10.13                   \\
\hline
Self-Attention Transformer           & 13.96              &    13.47 \\
\hline
\end{tabular}
\end{table*}


\begin{figure}[ht] 
  \label{ fig6} 
  \begin{minipage}[b]{0.5\linewidth}
    \centering
    \includegraphics[width=0.8\textwidth]{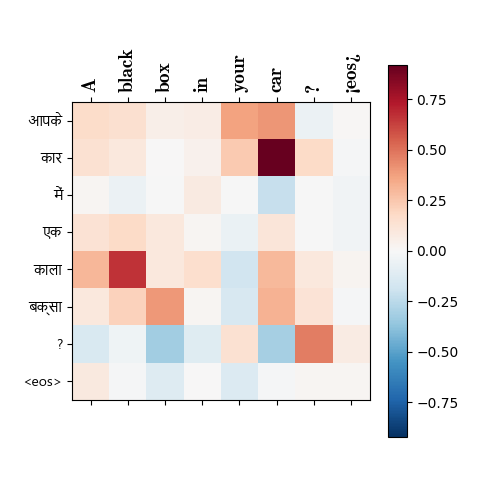}
  \subcaption{Sample 1}
 \vspace{4ex}
  \end{minipage}
  \begin{minipage}[b]{0.5\linewidth}
    \centering
    \includegraphics[width=1.0\textwidth]{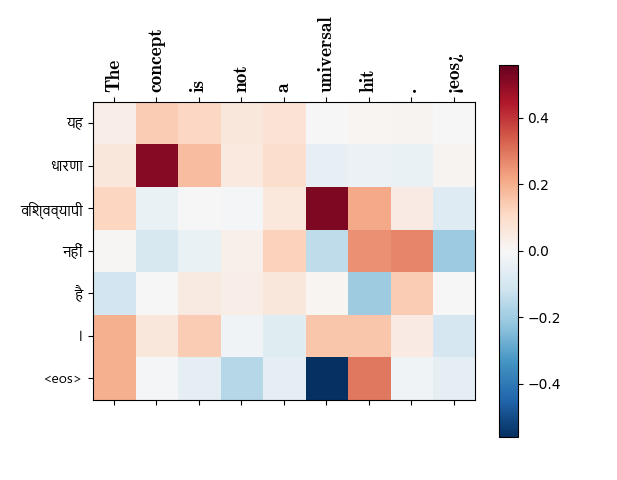} 
  \subcaption{Sample 2}
    \vspace{4ex}
  \end{minipage}
  \caption{Hindi-English Heatmap}
  \label{fig:en-hi}
  \end{figure}
  
  \begin{figure}[ht] 
  \label{ fig7} 
  \begin{minipage}[b]{0.5\linewidth}
    \centering
    \includegraphics[width=1.0\textwidth]{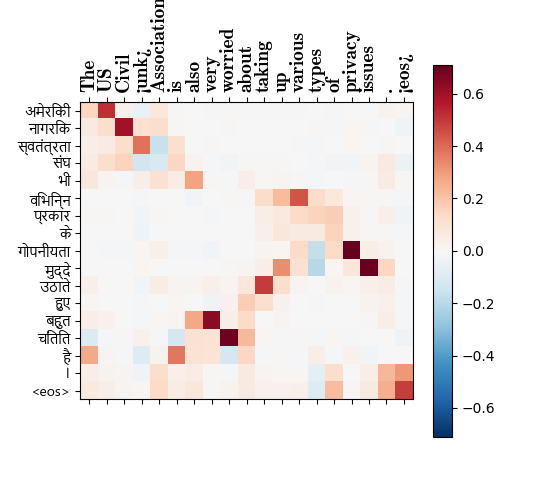} 
  \subcaption{Sample 1}
    \vspace{4ex}
  \end{minipage}
  \begin{minipage}[b]{0.51\linewidth}
    \centering
    \includegraphics[width=0.8\textwidth]{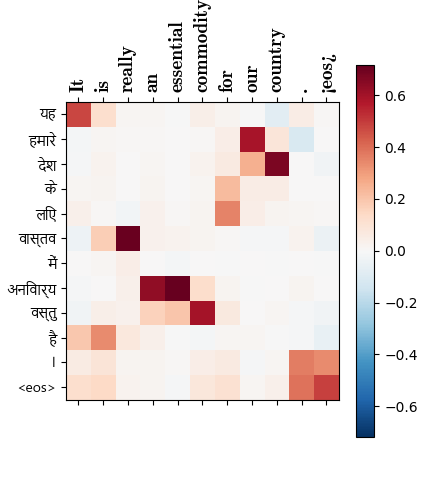}
  \subcaption{Sample 2}
    \vspace{4ex}
  \end{minipage} 
  \caption{English-Hindi Heatmap sample}
  \label{fig:hi-en}
\end{figure}

Figure \ref{fig:en-hi} shows the word-word association heat map for selected translated and target sentences when transformer model is trained on English-Hindi translation task and similarly Figure \ref{fig:hi-en} shows the word-word association heat map for selected translated and target sentences when transformer model is trained on Hindi-English translation task.

\section{Conclusion}

In this paper, we initially discussed about Machine translation. We started our discussion from a brief discussion on basic Machine translation objective and terminologies along with early Statistical approaches (SMT). We then discussed the role of deep learning models in improving different components of SMT, Then we shifted our discussion on end-to-end neural machine translation (NMT). Our discussion was largely based on the basic encoder-decoder based NMT, attention based model. We finally listed the challenges in Neural Translation models and mentioned future fields of study and open ended problems. Later we proposed a self-attention transformer network for Hindi-English language translation and compare this model with other neural machine translation models on the basis of BLEU. We concluded our study by delineating the advantages and disadvantages of all the three models.

\bibliographystyle{unsrt}  
\bibliography{references}  


\end{document}